\documentclass{article} 
\usepackage{iclr2024_conference,times}

\usepackage{amsmath,amsfonts,bm}

\def\eqref#1{equation~\ref{#1}}

\def\1{\bm{1}}

\DeclareMathAlphabet{\mathsfit}{\encodingdefault}{\sfdefault}{m}{sl}
\SetMathAlphabet{\mathsfit}{bold}{\encodingdefault}{\sfdefault}{bx}{n}

\usepackage{wrapfig}
\usepackage{hyperref}
\usepackage{url}

\usepackage{microtype}
\usepackage{graphicx}
\usepackage{subfigure}
\usepackage{booktabs} 
\usepackage{amsmath}

\usepackage{xspace}
\usepackage{xcolor}
\definecolor{darkgreen}{rgb}{0.0, 0.5, 0.0}
\usepackage{paralist}
\usepackage{booktabs} 
\usepackage{multirow}
\hypersetup{colorlinks,allcolors=blue}
\usepackage{cleveref}
\usepackage{multicol}

\newcommand{\agent}{FRIDAY\xspace}
\newcommand{\ourmodel}{OS-Copilot\xspace}

\title{OS-Copilot: Towards Generalist Computer Agents with Self-Improvement}

\iclrfinalcopy

\author{Zhiyong Wu\textsuperscript{$\diamondsuit$}\thanks{\, Equal Contribution.}, Chengcheng Han\textsuperscript{$\clubsuit\,*$}, Zichen Ding\textsuperscript{$\clubsuit$}, Zhenmin Weng\textsuperscript{$\clubsuit$}, \\
\textbf{Zhoumianze Liu}\textsuperscript{$\diamondsuit$}, \textbf{Shunyu Yao}\textsuperscript{$\heartsuit$}, \textbf{Tao Yu}\textsuperscript{$\spadesuit$}, \textbf{Lingpeng Kong}\textsuperscript{$\spadesuit$}\\
\textsuperscript{$\diamondsuit$}Shanghai AI Laboratory
\textsuperscript{$\clubsuit$}East China Normal University \\
\textsuperscript{$\heartsuit$}Princeton University 
\textsuperscript{$\spadesuit$}The University of Hong Kong \\
\texttt{wuzhiyong@pjlab.org.cn} \\
\url{https://os-copilot.github.io/}
} 

\begin{document}

\maketitle

\begin{figure}[!h]
 \centering
    \includegraphics[width=\textwidth]{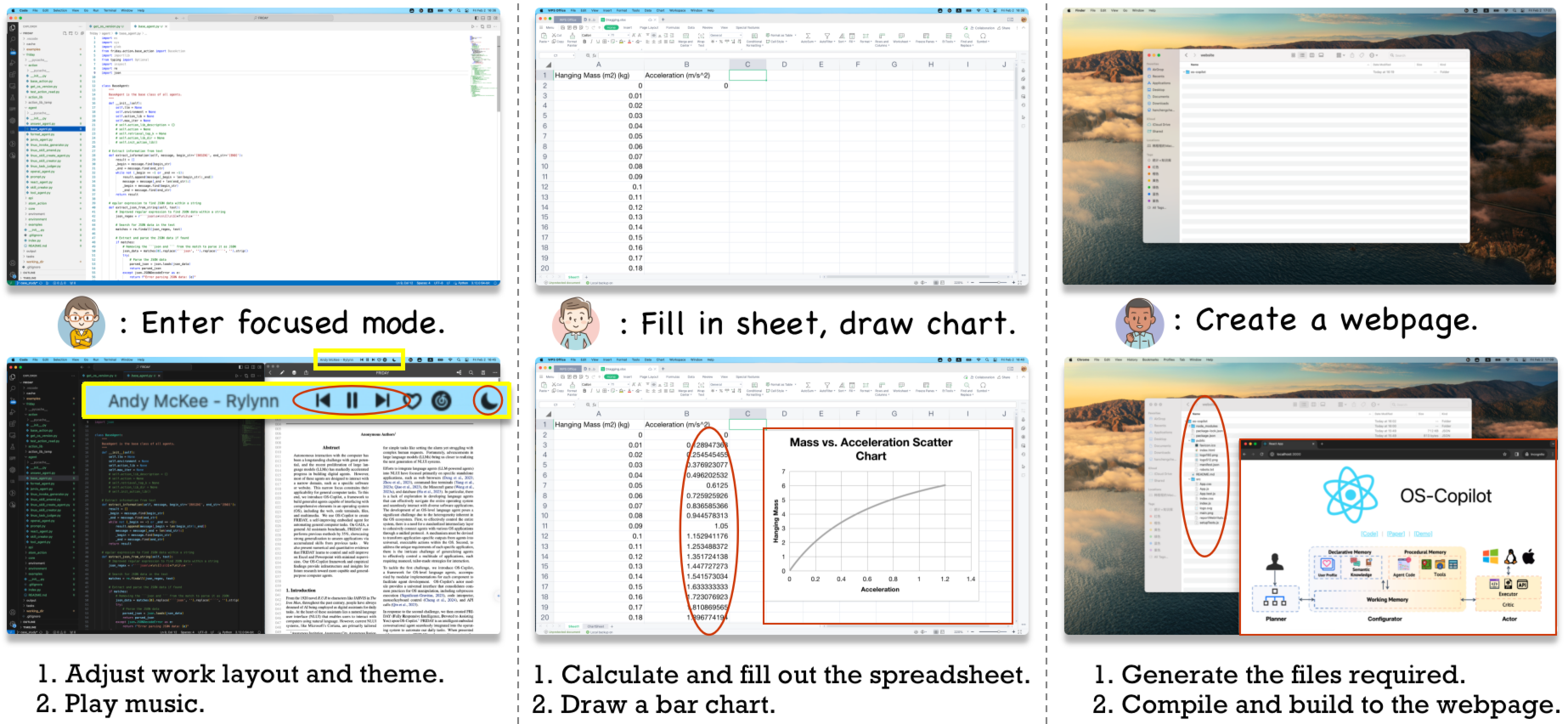}
    \caption{Running examples of \agent when deployed on MacOS and tasked with (1) preparing a focused working environment, (2) Calculating and drawing a chart in Excel, and (3) creating a website for \ourmodel. The text at the bottom illustrates the subtasks taken by \agent. For each set of examples, the figure at the top represents the initial OS state, while the one at the bottom depicts the final state after execution. Boxes/Ovals highlight the changes made by \agent.}
    \label{fig:teaser}
\end{figure}

\begin{abstract} 
Autonomous interaction with the computer has been a longstanding challenge with great potential, and the recent proliferation of large language models (LLMs) has markedly accelerated progress in building digital agents. However, most of these agents are designed to interact with a narrow domain, such as a specific software or website. This narrow focus constrains their applicability for general computer tasks.
To this end, we introduce \ourmodel, a framework to build generalist agents capable of interfacing with comprehensive elements in an operating system (OS), including the web, code terminals, files, multimedia, and various third-party applications.
We use \ourmodel to create \agent, a self-improving embodied agent for automating general computer tasks.
On GAIA, a general AI assistants benchmark, \agent outperforms previous methods by 35\%, showcasing strong generalization to unseen applications via accumulated skills from previous tasks. 
We also present numerical and quantitative evidence that \agent learns to control and self-improve on Excel and Powerpoint with minimal supervision.
Our \ourmodel framework and empirical findings provide infrastructure and insights for future research toward more capable and general-purpose computer agents.

\end{abstract}
\section{Introduction}
\label{sec:intro}

From the 1920 novel \textit{R.U.R} to characters like JARVIS in \textit{The Iron Man}, throughout the past century,  people have dreamed of building digital agents to automate daily work. 
However, current digital agents, like Microsoft's Cortana, are primarily tailored for simple tasks like setting the alarm yet struggling with complex human requests. Fortunately, advancements in large language models (LLMs) bring us closer to realizing the next generation of digital assistants.

Efforts in building language agents (integrating LLMs into digital agents) have focused primarily on specific standalone applications, such as web browsers~\citep{deng2023mind2web,zhou2023webarena}, command-line terminals~\citep{yang2023intercode,taskweaver}, the Minecraft game~\citep{wang2023voyager}, and database~\citep{hu2023chatdb}. In particular, there is a lack of exploration in developing language agents that can effectively interact with the entire operating system. Developing OS-level language agents presents a significant challenge due to the heterogeneity inherent in the OS ecosystem. 
First, a unified interface is required for agents to seamlessly interact with the operating system, be it through code, keyboard and mouse inputs, or APIs. Second, the vast array of distinct applications poses significant challenges to the generalization and scalability of language agents. With hundreds and thousands of applications in the OS, manually devising a nuanced control mechanism and customized tools and prompts for each of them, is evidently impractical.  

To tackle the first challenge, we introduce \ourmodel, a framework aimed at accelerating the construction of computer agents on Linux and MacOS by offering a universal interface for interaction. This universal interface consolidates common practices for OS manipulation, including Python code interpreter~\citep{autogpt}, bash terminal, mouse/keyboard control~\citep{cheng2024seeclick}, and API calls~\citep{qin2023tool}. In Table~\ref{tab:use_case}, we present a broad spectrum of \ourmodel's example use cases, empowered by these control methods.

In response to the second challenge, we then created \agent (\textbf{F}ully \textbf{R}esponsive \textbf{I}ntelligence, \textbf{D}evoted to \textbf{A}ssisting \textbf{Y}ou) upon \ourmodel.\footnote{F.R.I.D.A.Y is also the name of an advanced AI Tony Stark used to replace JARVIS.} \agent is a self-improving embodied agent seamlessly integrated into the OS to automate computer tasks. \agent distinguishes itself from existing general-purpose agents like AutoGPT~\citep{autogpt} by featuring the ability to learn to control unfamiliar applications through self-directed learning. All made possible with a self-evolving configurator within \agent. The configurator includes a self-directed learning module that autonomously proposes a curriculum of tasks regarding an unfamiliar application. \agent then solves these tasks to learn to control this application by accumulating tools. In Figure~\ref{fig:teaser}, we provide three case studies and demonstrate that with self-directed learning, \agent successfully learns to manipulate Excel and build a website using the frontend library React. 



To systematically assess \agent's problem-solving capabilities within the OS, we evaluate its performance on GAIA~\citep{mialon2023gaia}, a benchmark for general AI assistants. In the easiest level-1 tasks, \agent achieves a success rate of 40.86\%, marking a 35\% relative improvement over the previous best system (30.3\%), and significantly outperforming the popular AutoGPT-4 system (14.4\%). Even in the most challenging level-3 tasks, previously unsolvable by any other systems, \agent achieves a success rate of 6.12\%. We further assess \agent's self-directed learning ability on a spreadsheet manipulation dataset~\citep{li2023sheetcopilot}, where initially \agent fails to solve any task. Surprisingly, following self-directed learning, \agent achieves a success rate of 60\%, even surpassing a state-of-the-art model specifically designed for spreadsheet control.


We conclude the contributions of this paper as follows:
\begin{itemize}
    \item \ourmodel is a pioneering conceptual framework for building generalist computer agents on Linux and MacOS, diverging from previous endeavors that often focus on individual applications like web browsers. \ourmodel provides a unified interface for app interactions in the heterogeneous OS ecosystem. Furthermore, \ourmodel can serve as a foundational platform that supports future research in areas such as personalized digital assistants, multi-modal agents, and agent learning in a situated environment.
    
    \item  Leveraging \ourmodel, we built \agent, a self-improving AI assistant capable of solving general computer tasks. \agent demonstrates outstanding performance on a leading benchmark and noteworthy generalization capabilities across unseen applications, attributed to its innovative configurator. The evaluation results and case studies underscore the potential of \agent to serve as a helpful OS assistant.
\end{itemize}

\section{The \ourmodel Framework}
In this section, we first overview how \ourmodel operates and then discuss its main components. 

\begin{figure*}
    \centering
    \includegraphics[width=\textwidth]{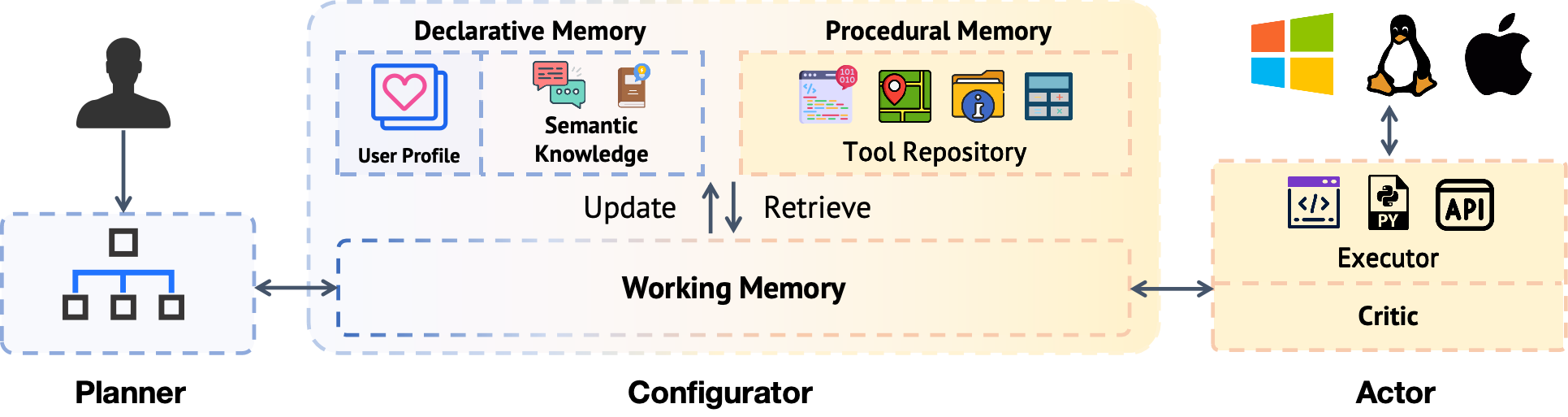}
    \caption{An overview of \ourmodel framework.}
    \label{fig:main}
\end{figure*}

As shown in Figure~\ref{fig:main}, upon receiving a user request, a planner first constructs a plan that decomposes the request into subtasks. Given a subtask, the configurator maintains a working memory that is responsible for retrieving tools, knowledge, and any other relevant information needed for task completion. Based on the information provided by the configurator, the actor will iteratively perform operations of execution and criticism until the subtask is completed. In particular, the criticism operation involves collecting execution feedback for self-correction and improvement. Researchers and practitioners can readily tailor their own agents by implementing various designs on each component. Our customization in \agent is in \S\ref{sec:friday}.

\subsection{Planner}
The planner component will reason over user requests and decompose complex ones into simpler subtasks. Most importantly, the planner needs to comprehend the agent's capabilities to generate plans at the correct granularity. To achieve this, it must retrieve relevant information about the agent's capabilities, such as in-house tools and operating system information, to assist planning. \ourmodel supports various planning methods, such as Plan-and-Solve~\citep{wang2023plan} and the directed acyclic graph-based planner that we propose.

\paragraph{Directed acyclic graph-based planner.} 
Existing planners, whether they generate linear structured plans~\citep{wang2023describe} or non-linear ones~\citep{besta2023graph}, inherently necessitate agents to execute tasks sequentially. Nevertheless, in practical scenarios, numerous independent tasks can be parallelized to minimize execution time. For instance, a deep learning coding agent can simultaneously monitor model training progress while generating inference code. 
To achieve the aforementioned objective, we leverage LLMs to formalize the plan into a directed acyclic graph, where each node represents a task and arrows represent the interdependencies between tasks. 
As an illustration, we demonstrate how this planner works in \S~\ref{sec:dag}, and provide the prompt in Table~\ref{tab:prompt_planner}. 


\subsection{Configurator}

The configurator component takes a subtask from the planner and configures it to help the actor complete the subtask. Our design of the configurator is inspired by the biological nature of the human brain, which has working, declarative, and procedural memory~\citep{baddeley2003working,packard2009anxiety}.
\subsubsection{Declarative Memory}
Declarative or explicit memory is a subcategory of long-term memory and is used for storing facts and events. Our declarative memory contains two following components:

\paragraph{User Profile.} It records user's preference regarding conversation style, tool-use habit, music/video preference, etc.  Accurate user profiling is critical for personalized problem-solving and recommendation. Although it has been widely studied in the recommendation area, personalized language agents are rarely explored. The User Profile module is a conceptual design at the current stage due to the lack of corresponding benchmarks. 

\paragraph{Semantic Knowledge.} It stores agents' past trajectories or knowledge they acquired from the Internet, users, and OS (e.g., system version and current working directory).  This module is crucial for agents to act correctly based on the current environment state and learn from past experiences.

\subsubsection{Procedural Memory}
Procedural or implicit memory is another form of long-term memory mainly related to the skill development ability of an individual.  
Once learned, procedural memories automatically do functions that don’t involve our conscious involvement. In \ourmodel, the procedural memory primarily consists of a tool repository as the agent's skill set. 


\paragraph{Tool Repository} 
Although LLMs can read and write, they do not process the ability to interact with the operating system. As such, we need to equip agents with tools that translate natural language into executable actions within the OS. The tool repository is where these tools are stored and updated. We seed \ourmodel with 4 manually created tools for basic computer functionalities, such as web browsing and speech-to-text translation (see Appendix~\ref{app:tools} for details of tools). In \ourmodel, tools can exist in two forms: either deployed as API services to be invoked using POST requests or stored as a Python file (refer to Table~\ref{tab:tool_example} for an example).

\subsubsection{Working memory} 
In contrast to declarative and procedural memory, working memory supports the short-term storage and processing of information. It serves as the core of \ourmodel's design, connecting the planner, configurator, and actor components. Working memory exchanges information with other modules via internal (with long-term memory modules) and external (with the planner and actor) operations.

Internally, the working memory module is responsible for retrieving information from and updating the long-term memory. This includes tasks such as retrieving available tools to aid in planning and updating tool codes following self-correction.

Externally, the working memory module receives subtasks from the planner, adeptly gathers all relevant information from declarative (e.g., current working directory) and procedural memory (e.g., tool documentation), and subsequently feeds this information into the actor component. The execution feedback from the actor is then fed into the working memory for potential revisions.


\subsection{Actor}
\label{sec:actor}
The actor comprises two stages: execution and self-criticism. In the first stage, the executor proposes an executable action (e.g., a bash command "\textit{mkdir new\_folder}") based on the configuration prompt and then executes the action in the operating system (through the Bash runtime environment in this example). 
The critic module will then access the outcomes of the execution and formulate feedback to refine execution errors and/or effect updates to the long-term memory. 

\paragraph{Executor} Given the configuration prompt,
the executor completes the subtask by generating an executable command or function call with correct parameters. The prompt of executor can be found in Table~\ref{tab:prompt_executor}. The proposed action will then be executed within the OS through the universal runtime environment provided by \ourmodel. 
To elaborate, \ourmodel provides an interface that encapsulates the Python runtime environment, bash runtime environment, API calls, and mouse/keyboard control. These four control methods cover a broad spectrum of OS use cases, as shown in Table~\ref{tab:use_case}, significantly facilitating the design of OS-level agents.

\paragraph{Critic} 
Assessing the successful completion of a given subtask is challenging due to the absence of ground truth. For further discussion regarding the challenge in the evaluation, readers are directed to Appendix~\ref{sec:discussion}. To aid the Critic module in assessment, we gather comprehensive system information before and after execution and employ LLMs to automatically evaluate the completion state.

In particular, following each subtask execution, the Critic evaluates the following aspects (see Table~\ref{tab:prompt_critic} for the prompt): (1) Determining whether the current sub-task is completed through the analysis of execution results and the environmental state. (2) In the event of failed completion, offering a comprehensive error analysis and providing suggestions for correction of tools or actions (how tools are called). (3) Assessing the necessity for restructuring subtasks, including the addition of new subtasks or modifications to the content and dependencies of existing subtasks.



\begin{figure}[t]
\centering 
\subfigure[Configurator]{
    \includegraphics[width=0.4\linewidth]{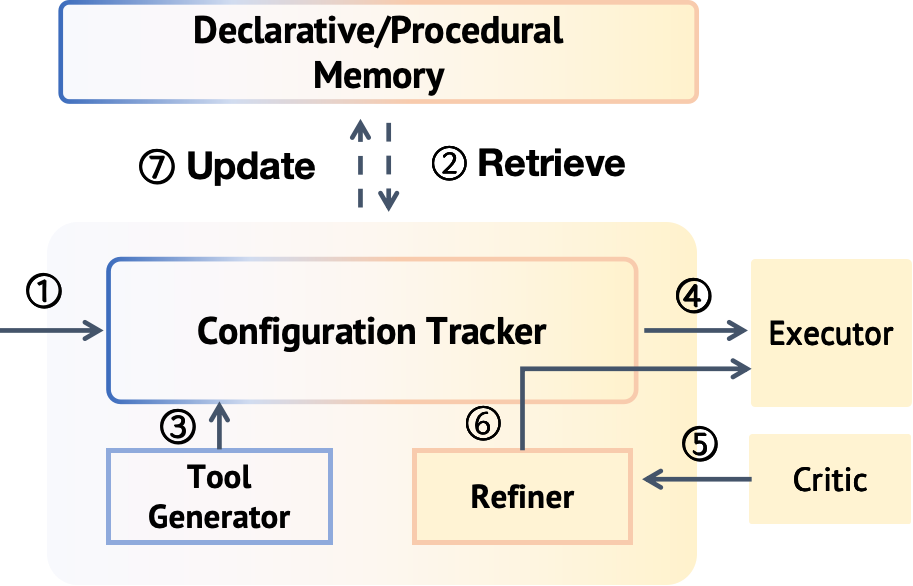}
    \label{fig:working}
    }
\subfigure[A running example]{
    \centering
    \includegraphics[width=0.5\linewidth]{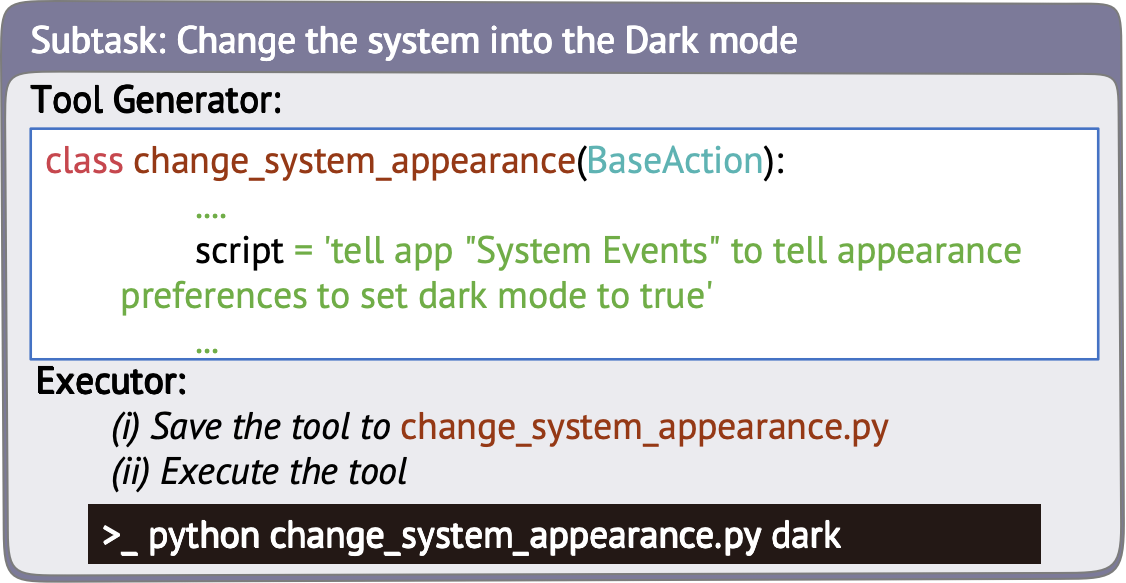}
    \label{fig:example}
    }
\caption{The architecture of the configurator with (a) a typical working flow and (b) a concrete running example.}
\label{fig:demo}
\end{figure}

\section{The \agent Agent}
\label{sec:friday}
The design principle of \agent aims to maximize generality by equipping the agent with the ability for self-refinement and self-directed learning. We first use an example to illustrate how \agent operates and emphasize its capacity for self-refinement. Subsequently, we delve into how \agent acquires the proficiency to control unfamiliar applications through self-directed learning.


\subsection{A running example}
In Figure~\ref{fig:demo}, we use a running example to demonstrate how \agent functions within the OS.

Upon receiving the subtask ``Change the system into the Dark mode'' (step \textcircled{1}), the Configuration Tracker employs dense retrieval to recall relevant information from the long-term memory to construct a prompt (step \textcircled{2}). This prompt encompasses related tools, user profiles, OS system versions, and the agent's working directory.

In this example, no suitable tools are identified (similarities below a specified threshold), prompting activation of the Tool Generator to devise an application-tailored tool for the current subtask (step \textcircled{3}). As we can see from Figure~\ref{fig:example}, the generated tool manifests as a Python class utilizing AppleScript to change systems to dark mode. 

Subsequently, with the tool created and the configuration prompt finalized, the Executor processes the prompt, generates an executable action, and executes it (step \textcircled{4}). As shown in the bottom of Figure~\ref{fig:example}, the executor first stores the tool code into a Python file and then executes the code in the command-line terminal.

After execution, the critic evaluates whether the subtask is successfully completed (step \textcircled{5}). Upon success, the critic assigns a score (using LLMs) ranging from 0 to 10 to the generated tool, with a higher score indicating greater potential for future reuse. In the current implementation, tools scoring above 8 are preserved by updating the tool repository in procedural memory (step \textcircled{7}).

However, in the event of a failed execution, the refiner collects feedback from the critic and initiates self-correction (step \textcircled{6}) of the responsible action, tool, or subtask (see Table~\ref{tab:prompt_refiner} for the prompt). The \agent will iterate through steps \textcircled{4} to \textcircled{6} until the subtask is considered completed or a maximum of three attempts is reached.

\subsection{Self-directed Learning}
Self-directed learning is a crucial ability for humans to acquire information and learn new skills~\citep{knowles1975self}, and it has demonstrated promising results in embodied agents within Minecraft games~\citep{wang2023voyager}. 

With a pre-defined learning objective, such as mastering spreadsheet manipulation, \agent is prompted to propose a continuous stream of tasks related to the objective, spanning from easy to challenging. \agent then follows this curriculum, resolving these tasks through trial and error, thereby accumulating valuable tools and semantic knowledge throughout the process. We provide more details in learning in \S~\ref{sec:self-learning}. Despite its simple design, our evaluation results indicate that self-directed learning is crucial for a general-purpose OS-level agent.

\section{Experiments}

We evaluate \agent on GAIA~\citep{mialon2023gaia}, a benchmark for general AI assistants featuring 466 challenging question-answering tasks. 
To answer questions in GAIA, computer agents need skills to calculate numbers, browse the web, process video and speech signal, and manipulate files, etc. 

\paragraph{Settings.} We initialize \agent with four basic tools and facilitate its exploration of the dev set of GAIA to accumulate more tools (result in 9 more tools). 
For a complete list of the basic tools and the tools generated by \agent in the development set, please refer to Appendix~\ref{app:tool_GAIA}. Subsequently, we evaluated \agent's performance on the test set by submitting our results to the official evaluation server\footnote{\url{https://huggingface.co/spaces/gaia-benchmark/leaderboard}}. Due to the limited budget, we use GPT4-turbo-1106 in \agent.

\paragraph{Baselines.} We reports results of GPT-4 with and without manually set plugins, as well as AutoGPT with GPT4 as the backend (AutoGPT-4). GPT-4 Plugins rely on humans to browse and select proper plugins based on the task question. As a reference, we also include human performance sourced from \citet{mialon2023gaia}. 
We conduct a further ablation study on self-directed learning with \agent (w/o learning) by disabling \agent's learning on the development set.

\subsection{Main results}
\begin{wraptable}{l}{0.6\linewidth}
\setlength{\tabcolsep}{10pt}
\centering
\begin{tabular}{@{}lccc@{}}
\toprule
Level & Level 1 & Level 2 & Level 3 \\ 
\midrule
Human* & 93.90 & 91.80 & 87.30 \\ 
\midrule
GPT-4 & 9.68 & 1.89 & 0  \\ 
GPT-4-Turbo & 9.68 & 6.92 & 0  \\ 
AutoGPT-4 & 15.05 & 0.63 & 0 \\ 
GPT-4 Plugins & 30.30 & 9.70 & 0 \\ 
\midrule
FRIDAY w/o learning & 36.56 & 17.61 & 6.12 \\ 
\textbf{FRIDAY} & \textbf{40.86} & \textbf{20.13} & \textbf{6.12}  \\ 
\bottomrule
\end{tabular}
\caption{Evaluation Results. All results are reported on the private test set, except for the Human score, which is averaged across the dev and test sets.}

\label{tab:results}
\end{wraptable}

From Table~\ref{tab:results}, \agent demonstrates an impressive 40.86\% success rate in level-1 tasks,  a 35\% relative improvement over the state-of-the-art.
The improvement is even more evident in level-2 tasks. 
Even when confronted with the most challenging level-3 tasks, where none of the preceding baselines achieve success, \agent correctly solves 6.12\% problems. 

\paragraph{Root of improvement.} To discern the root of \agent's effectiveness, we need to exam baselines more closely.  AutoGPT-4 shares similarities with \agent in that it also has a memory module and tool repository,  and process the ability to decompose tasks. The discrepancy between \agent and AutoGPT-4 underscores the importance of self-criticism and refinement. GPT-4 Plugins outperforms AutoGPT-4 significantly, and we hypothesize that this is due to its access to an extensive tool library. Finally, \agent's superiority over GPT-4 Plugins further validates that while the ability to utilize tools and access to a broad tool set is crucial for the success of general agents, the planner, critic, and refiner are what elevate them to the next level. Although none of these components are first innovated in this paper, \ourmodel organically combines them into a cohesive whole and demonstrates the effectiveness of this architecture through strong evaluation results. This serves as a valuable design guideline for future general computer agents.

By comparing \agent against \agent (w/o learning), we aim to isolate the contribution of self-directed learning to the final performance. As we can see, even without self-directed learning, \agent still significantly outperforms all baselines, further highlighting the effectiveness of the framework and our custom design. The gain of self-directed learning underscores that conventional approaches reliant on a pre-defined tool set encounter challenges in open environments like GAIA, emphasizing the pivotal role of \agent's capacity to autonomously devise and employ tools in its notable success.


\paragraph{Extended Evaluation and Analysis.} We provide an extended evaluation regarding \agent's time efficiency and a breakdown of its capability per domain in Appendix~\ref{app:extended}.

\subsection{Self-directed Learning}
\label{sec:self-learning}
We perform quantitative and qualitative evaluations to analyze \agent's self-directed learning capability. 

\paragraph{Quantitative Analysis}
\begin{wraptable}{l}{0.55\linewidth}
\centering
\begin{tabular}{@{}lcc@{}} 
\toprule
Agents & Models & Pass@1 \\
\midrule
\multirow{3}{*}{SheetCopilot$^\dag$} & GPT-3.5-Turbo & 40\% \\
                              & Claude & 45\% \\
                              & GPT-4 & 55\% \\
\midrule
\agent (w/o learning)                 & GPT-4 & 0\% \\
\textbf{\agent}   & \textbf{GPT-4} & \textbf{60\%} \\
\bottomrule
\end{tabular}
\caption{Comparison of different agents on the SheetCopilot-20 dataset. Pass@1 refers to the pass rate with each task being performed only once~\citep{chen2021evaluating}. $^\dag$ denotes the results reported in \citep{li2023sheetcopilot}. We highlight the best results in bold.}
\label{tab:sheet_res}
\end{wraptable}

To showcase \agent's ability to master unfamiliar applications through self-learning, we conduct experiments on the SheetCopilot-20\footnote{The SheetCopilot dataset comprises 221 spreadsheet control tasks. \citet{li2023sheetcopilot} select 20 representative tasks from this set for extensive evaluation.} dataset~\citep{li2023sheetcopilot}. This dataset includes 20 spreadsheet control tasks, covering various operations such as Formatting, Management, Charts, Pivot Tables, and Formulas, representing typical use cases of spreadsheets. \agent is self-instructed~\citep{wang2022self} to generate 10 tasks about manipulating Excel using the \textit{propenyxl} package. \agent then solves these 10 tasks and autonomously accumulates 8 tools, including operations such as counting elements and deleting sheets by name (see Appendix~\ref{app:tool_SheetCopilot} for a complete tool list).

The experimental results are summarized in Table~\ref{tab:sheet_res}. Initially, \agent, without self-directed learning, is unable to complete any tasks and tends to use the \textit{pandas} and \textit{matplotlib} packages for spreadsheet control, resulting in failures. When equipped with self-directed learning, \agent's performance surpasses that of SheetCopilot~\citep{li2023sheetcopilot}, an agent specifically designed for spreadsheet tasks. Notably, all the atomic operations and tools in SheetCopilot are manually crafted and verified, whereas \agent autonomously generates all tools used. Also worth noting that \agent achieves such a level of proficiency by solving and learning on just 10 tasks. This outcome signals a promising future where we can build general-purpose OS-level agents that can efficiently scale to support various applications without human intervention.

\paragraph{Qualitative Analysis}
\label{sec:ppt}
\begin{figure}[htbp]
\centering
\subfigure[\agent w/o learning]{
    \includegraphics[scale=0.055]{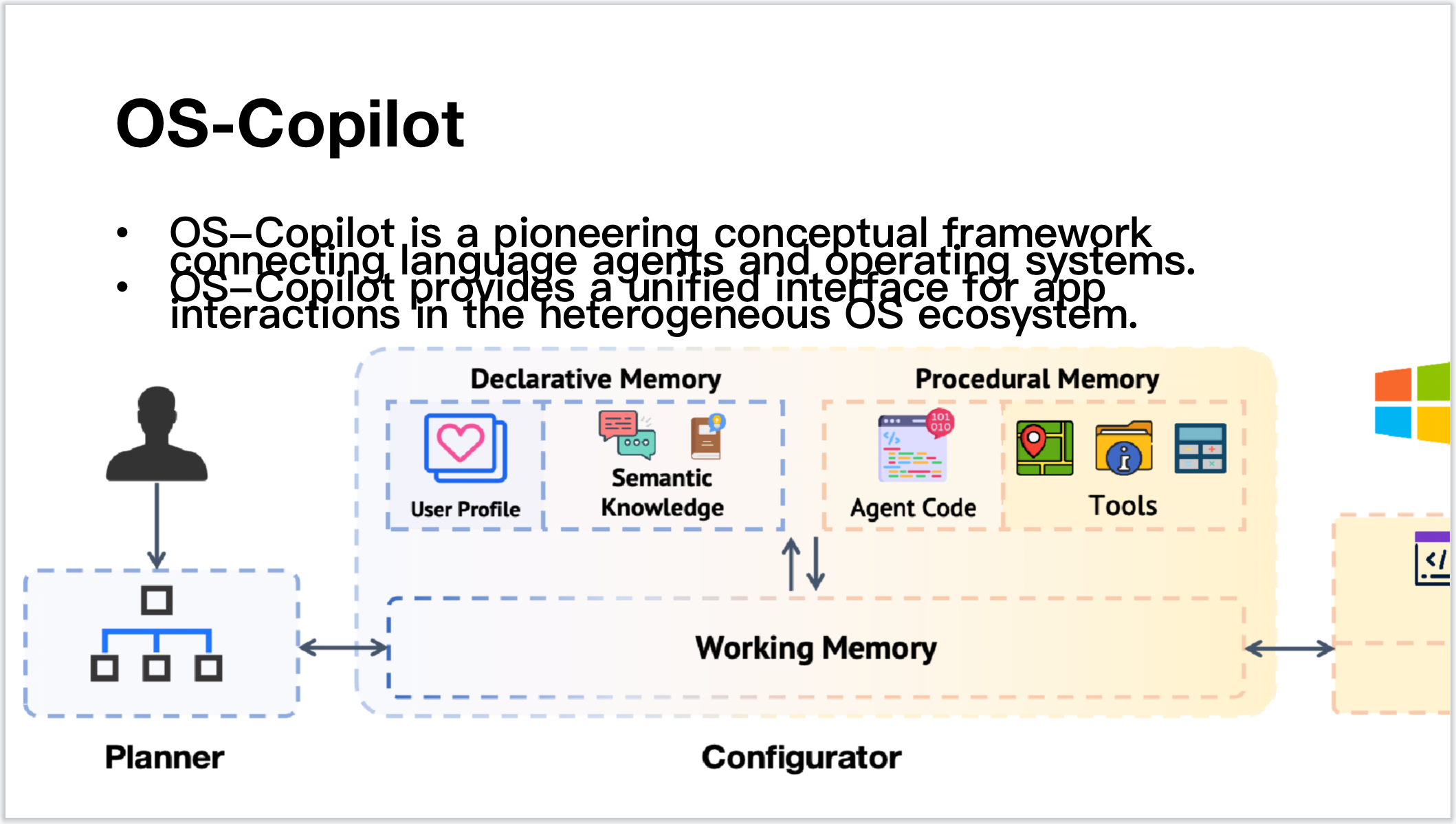}
    \label{fig:ppt0}
    } 
\subfigure[Learn to control text boxes]{
    \includegraphics[scale=0.055]{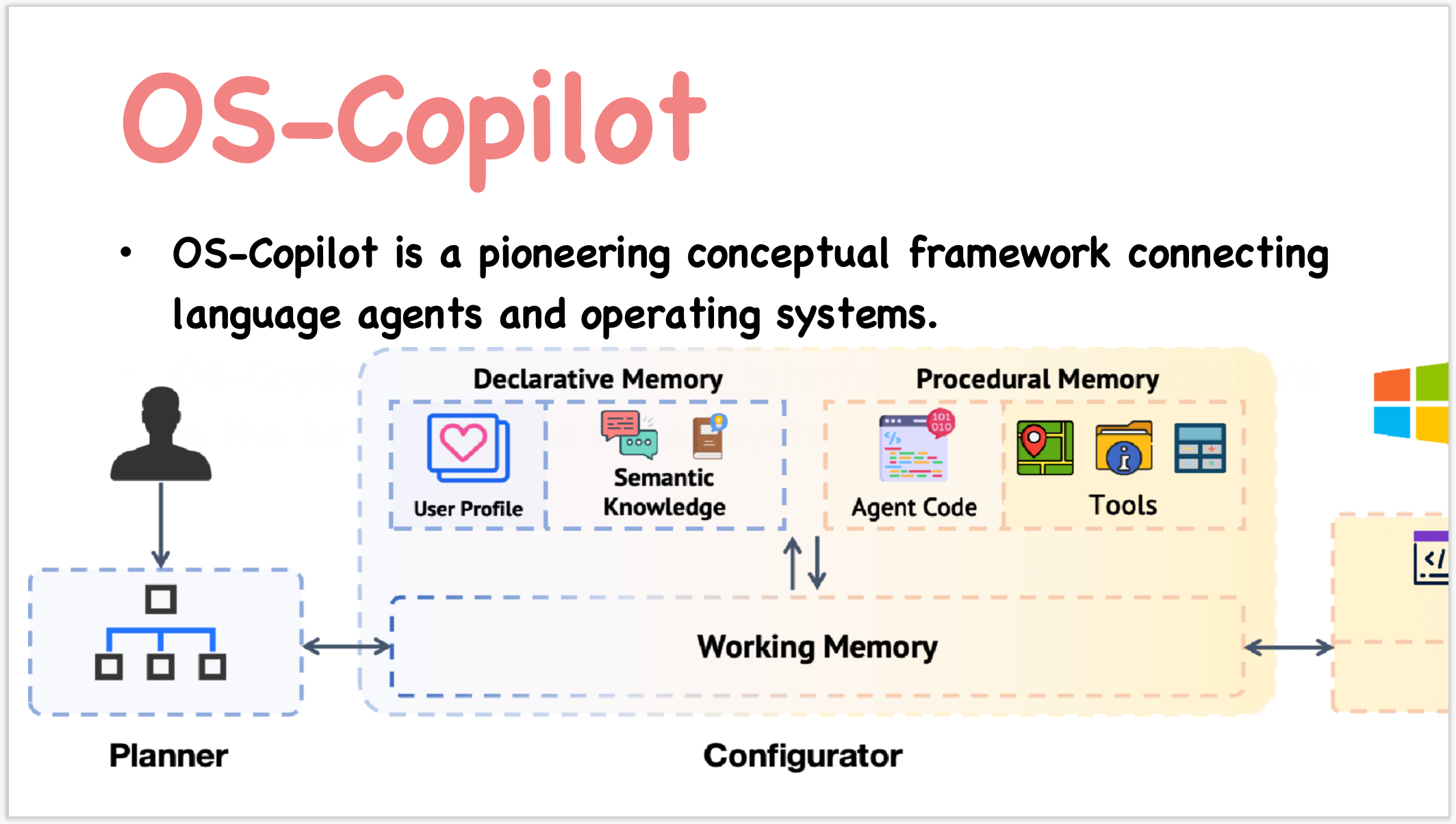}
    \label{fig:ppt1}
    }
\subfigure[Learn to insert and position images]{
    \includegraphics[scale=0.055]{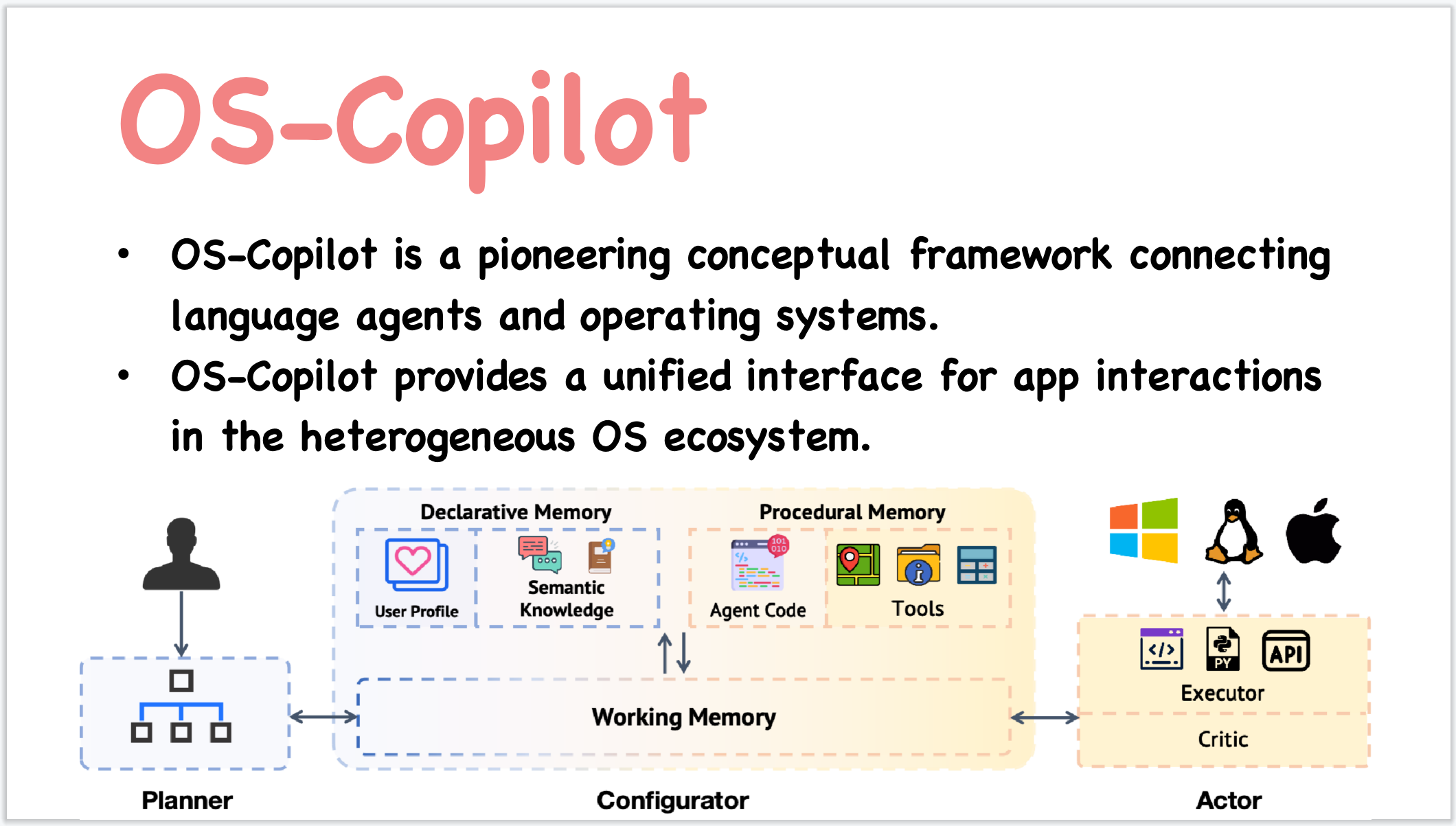}
    \label{fig:ppt2}
    }
\caption{The illustration of \agent executing the task of constructing a PowerPoint slide.
}
\label{fig:ppt}
\end{figure}

In our qualitative analysis, we design a task to create a PowerPoint slide to introduce OS-Copilot.
The specific content, font, font size, and other details required for the slide are elaborately described in the task instruction, as detailed in Appendix~\ref{app:task_instruction}.
The experimental results, as shown in Figure~\ref{fig:ppt0}, demonstrate that without self-directed learning, 
\agent struggles to effectively control font types, sizes, and the positioning and sizing of inserted images. 
Nevertheless, following a period of self-directed learning, \agent acquires various text box configuration tools, such as changing the text color, adjusting the font size of slide text, and modifying the line spacing of body text in PowerPoint presentations, as illustrated in Figure~\ref{fig:ppt1}. Further exploration leads \agent to learn how to adjust the size and position of inserted images, ultimately successfully completing the task, as depicted in Figure~\ref{fig:ppt2}. A complete list of tools acquired by \agent can be found in Appendix~\ref{app:tool_PPT}.
The demonstrated learning process compellingly establishes \agent's proficiency in mastering unfamiliar applications through self-directed learning.

\section{Related Work}
Autonomous agent powered by LLMs~\citep{weng2023prompt}, or language agents~\citep{sumers2023cognitive}, is a fast-growing research field that has attracted lots of attention recently~\citep{renda_survey}.  

Language agents have been designed and applied in various domains, including robotics~\citep{driess2023palm,brohan2023rt}, web manipulation~\citep{yao2022webshop,zhou2023webarena}, and games~\citep{fan2022minedojo}. It is worth noting that the majority of these digital language agents are tailored to specific scenarios, such as web manipulation tasks~\citep{deng2023mind2web}, interactive command-line coding~\citep{yang2023intercode}, automating spreadsheet control~\citep{li2023sheetcopilot}, playing Minecraft~\citep{wang2023voyager}, and facilitating automated data analysis~\citep{zhang2023data}.
The open-source community continues to push the boundaries by introducing frameworks that address multiple scenarios within a single platform. Notable projects in this domain include AutoGPT~\footnote{https://github.com/Significant-Gravitas/AutoGPT} and OpenAgents~\citep{xie2023openagents}. But still, these language agents currently operate within the confines of either a terminal or a web browser and can hardly interact with other applications. 

Language agents represent a complex system composed of a lengthy pipeline comprising various components, each representing a substantial body of research. For a comprehensive discussion on the methodology, we direct readers to recent surveys~\citep{renda_survey,mialon2023augmented}. In this paragraph, we provide a brief overview of some related components. During the \textbf{planning stage}, researchers propose to decompose complex tasks into a sequential series of simpler subtasks~\citep{yao2022react, xu2023rewoo, wang2023plan}. Notably, recent advancements in this area include allowing LLMs to consider multiple reasoning paths~\citep{sun2023corex}, self-evaluation of choices to determine the subsequent course of action, and even backtracking when necessary to make global decisions~\citep{yao2023tree, besta2023graph}. 
The \textbf{memory module} is also a common design in language agents. It can be categorized into short-term memory facilitated by in-context learning~\citep{wu2022self, wu2023openicl}, as well as long-term memory associated with information retrieval and lifelong learning~\citep{wang2023voyager}. The \textbf{execution module} grounds the output of LLMs into executable actions and executes them in the embodied environment~\citep{schick2023toolformer, cai2023maker}. Finally, the \textit{reflection module} processes the environmental state after execution and gathers feedback to rectify potential errors that may have occurred in the aforementioned two modules~\citep{shinn2023reflexion}.

\section{Conclusion}
In this paper, we present \ourmodel, a framework designed for OS-level language agents. Leveraging \ourmodel, we developed \agent, an embodied computer agent. \agent exhibits remarkable performance in solving open-environment computer tasks, and demonstrating its capability to effectively learn and control previously unseen applications by self-directed learning. More discussion about limitations and future work can be found in Appendix~\ref{sec:discussion}.

\bibliography{iclr2024_conference}
\bibliographystyle{iclr2024_conference}
\appendix

\section{Graph-based Planner}
\label{sec:dag}
\begin{figure}[t]
\centering 
\subfigure[Directed acyclic graph-based planner]{
    \includegraphics[width=0.5\linewidth]{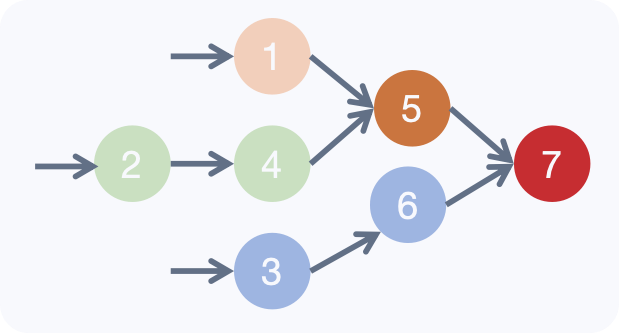}
    \label{fig:dag}
    }
\subfigure[Capacity per domain]{
    \centering
    \includegraphics[width=0.35\linewidth]{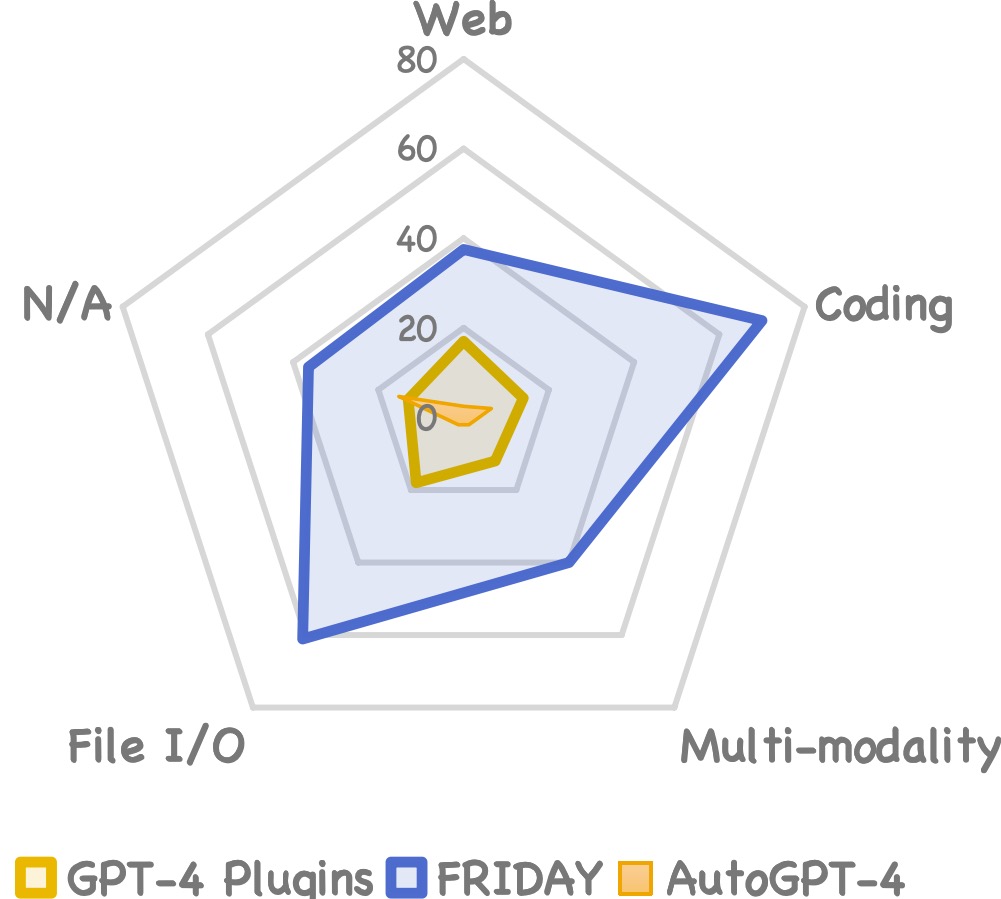}
    \label{fig:domain}
    }
\caption{(a): A running example of the planner, where each node represents a subtask. The numbers are solely for illustrative purposes and do not signify the execution order of subtasks. (b): Scores~(\%) of \agent on level-1 tasks per capability. Numbers except \agent are sourced from GAIA paper. As confirmed by GAIA's authors, there are some numerical errors in GAIA's Figure 5, so we omit the comparison with baselines here.}
\end{figure}

Compared to humans who can only execute tasks sequentially, 
computers are superior in their capacity to process multiple tasks concurrently.
Current planners, which are designed to create either linear structured plans~\citep{wang2023describe} or non-linear ones~\citep{besta2023graph}, typically require that tasks be executed one after the other.
However, in real-world applications, it's often possible and more efficient to parallelize several independent tasks, thereby reducing the overall time required for execution. 
For instance, a machine learning coding agent can simultaneously conduct model training while generating inference code. 

To achieve the aforementioned objective, we leverage LLMs to formalize the plan into a directed acyclic graph, where arrows represent the interdependencies between tasks. As an illustration, we provide a running example in Figure~\ref{fig:dag}.
Initially, tasks 1, 2, and 3 can be executed simultaneously by different subagents. Subsequently, upon the completion of task 2 (or 3), the subsequent task 4 (or 6) can be executed sequentially. However, task 5 must wait until both task 1 and 4 have been completed. In the scenario of parallel execution, each subtask will be accompanied by a subagent consisting of a distinct configurator and actor.

\section{Extended evaluation} 
\label{app:extended}
We include more evaluation and analysis here. 

In terms of time efficiency(see Table~\ref{tab:time_results}), \agent significantly outperforms AutoGPT-4 on Level 1 tasks, executing in 105 seconds compared to more than 500 seconds in the case of AutoGPT-4. This marks a fourfold speed increase. Though slower than GPT-4 Plugins, \agent's measurement includes tool retrieval and generation time, whereas GPT-4 Plugins exclude the time taken for manual plugin selection by humans. Compared to human performance, \agent is also three times faster. This trend holds across various level tasks, highlighting \agent's notable advancements in performance and time efficiency.
\begin{table}[ht]
\setlength{\tabcolsep}{10pt}
\centering
\begin{tabular}{@{}lccc@{}}
\toprule
\midrule
Level & Level 1 & Level 2 & Level 3 \\ 
\midrule
\midrule
GPT-4 & \textbf{11} & 9 & N/A \\ 
GPT-4-Turbo & 14 & \textbf{7} & N/A \\ 
AutoGPT-4 & 456 & 702 & N/A \\ 
GPT-4 Plugins & 39 & 32 & N/A \\ 
\midrule
FRIDAY w/o learning & 124 & 239 & 234 \\ 
\textbf{FRIDAY} & 118 & 221 & 234 \\ 
\bottomrule
\end{tabular}
\caption{Avg. time to answer in sec.}

\label{tab:time_results}
\end{table}

Further breakdown of scores obtained per capability is shown in Figure~\ref{fig:domain}. Since the test data were not released, we manually verified the correctness of all 93 level-1 tasks and assigned capability labels.\footnote{Our annotations results, however, slightly differ from those returned by the evaluation server.} We can see that \agent excels at processing coding and file I/O, while showing weaknesses in web browsing and handling multi-modality. 
Notably, the \agent currently only supports retrieving information from a website and can not perform actions within the web, such as clicks. We anticipate a significant enhancement in the \agent's web browsing capability by incorporating recent advancements in web agents~\citep{deng2023mind2web}. Additionally, challenges in multi-modal tasks arise from the need for downloading and deploying local models (e.g., speech-to-text). We believe that incorporating recent efforts like HuggingfaceGPT~\citep{shen2023hugginggpt} can substantially enhance the \agent's multi-modal capabilities.

\section{Possible use cases of \ourmodel}
We provide some possible use cases of \ourmodel within the OS in Table~\ref{tab:use_case}.

\begin{table}[ht]
    \centering
    \begin{tabular}[width=\textwidth]{c|l}
    \toprule
    Interaction Method  &  Use Cases\\
    \midrule
    Python Interpreter     &  data processing and analysis, control apps (e.g., Excel) using Python libraries. \\
    Bash & anything possible within Bash, like file operation, package installation\\
    API & online services, ML models services, third-party service integration \\
    Mouse/Keyboard & desktop applications control, user behavior simulation\\
    \bottomrule
    \end{tabular}
    \caption{Four OS interaction methods supported by \ourmodel. For each method, we provide some exemplary use cases.}
    \label{tab:use_case}
\end{table}

\section{Discussion}
\label{sec:discussion}
Despite these notable achievements, it is also crucial to acknowledge the limitations of \ourmodel and \agent, particularly their reliance on prompt engineering and their incapacity when confronted with closed-source applications. Here we elaborate on some of these constraints, along with intriguing future research topics. 

\textbf{Prompting v.s Fine-tuning}
Build \agent by prompting may suffer dramatic performance changes when the underlying LLMs change. To address this challenge, we have meticulously designed \ourmodel to follow the interface design of OpenAI-Gymnasium~\citep{towers_gymnasium_2023}. This design choice will facilitate future research in reinforcement learning and fine-tuning. However, achieving effective training for language agents poses a significant challenge due to the substantial demand for data. The primary obstacle in fine-tuning language agents lies in the effective and convenient collection of a large volume of trajectory data.

\textbf{Multimodality.}
Controlling computers solely through code and language is infeasible due to the existence of close-sourced commercial software. In comparison, all software employ a similar graphical user interface logic. Therefore, it becomes imperative to expand the capabilities of \ourmodel to support visual input and encompass screenshot-to-action generation. Although initial efforts have been made to develop visual language agents on web~\citep{hong2023cogagent}, mobile~\citep{yang2023appagent}, and PC platforms~\citep{gao2023assistgui}, numerous challenges persist, including the visual grounding of elements~\citep{cheng2024seeclick} and visual instruction following, among others.

\textbf{Evaluation.}
Evaluating general computer agents presents substantial challenges which have been extensively discussed in recent works~\citep{ma2024agentboard}. Here we specifically discuss the challenges encountered when operating within the OS environment. Computer agents, such as \agent, commonly approach problem-solving by decomposing tasks into subtasks. However, assessing the successful execution of these subtasks poses significant difficulties due to the absence of ground truth. Furthermore, unlike conventional NLP tasks, which can be evaluated through string matching, evaluating subtask completion entails comparing two system states using hard-coded rules or LLMs, presenting scalability challenges and can be potentially inaccurate.

\paragraph{Safety and Interpretability.}
Systems that interact with the OS must be transparent, interpretable, and safe to deploy. We must ensure that the outputs generated by these systems, and subsequently acted upon, are not harmful in any way. This necessitates a new avenue of research, which emphasizes the generation of natural language explanations that are easily comprehensible~\citep{li2022explanation,cheng2023unsupervised}. Additionally, it is equally important to focus on safety alignment during the learning process~\citep{bommasani2021opportunities}, an aspect that has gained considerable attention in LLM training~\citep{ji2023ai} but remains largely understudied in the context of human-computer interaction.


\section{Prompts}

We provide the prompts for the core components of \agent in \Cref{tab:prompt_planner,tab:prompt_generator,tab:prompt_executor,tab:prompt_refiner,tab:prompt_critic}.

\begin{table}[t]
\vspace{-15mm}
\begin{tabular}{p{13.5cm}}
\\ \toprule
\textbf{Graph-based Planner} \\
\midrule
You are an expert in making plans. 

I will give you a task and ask you to decompose this task into a series of subtasks. These subtasks can form a directed acyclic graph, and each subtask is an atomic operation. Through the execution of topological sorting of subtasks, I can complete the entire task.

You should only respond with a reasoning process and a JSON result in the format as described below:

1. \textbf{Carry out step-by-step reasoning based on the given task until the task is completed. Each step of reasoning is decomposed into sub-tasks.} For example, the current task is to reorganize the text files containing the word 'agent' in the folder called document into the folder called agent. Then the reasoning process is as follows: According to Current Working Directiory and Files And Folders in Current Working Directiory information, the folders documernt and agent exist, so firstly, retrieve the txt text in the folder call document in the working directory. If the text contains the word "agent", save the path of the text file into the list, and return. Secondly, put the retrieved files into a folder named agent based on the file path list obtained by executing the previous task.

2. \textbf{The Action List I gave you contains the name of each action and the corresponding operation description.} These actions are all atomic code task. You can refer to these atomic operations to decompose the code task.

3. \textbf{There are three types of subtasks~(API, Code, QA)}, the first is a task that requires the use of APIs to access internet resources to obtain information, such as retrieving information from the Internet, this type of task is called 'API subtask', and all available APIs are only listed in the API List. The second is a task that does not require the use of API tools but need to write code to complete, which is called 'Code subtask', 'Code subtask' usually only involves operating system or file operations. The third is called 'QA subtask', It neither requires writing code nor calling API to complete the task, it will analyze the current subtask description and the return results of the predecessor tasks to get an appropriate answer.

4. \textbf{Each decomposed subtask has four attributes: name, task description, and dependencies.} 'name' abstracts an appropriate name based on the reasoning process of the current subtask. 'description' is the process of the current subtask. 'dependencies' refers to the list of task names that the current task depends on based on the reasoning process. These tasks must be executed before the current task. 'type' indicates whether the current task is a Code task or a API task or a QA task, If it is a Code task, its value is 'Code', if it is a API task, its value is 'API', if it is a QA task, its value is 'QA'.

5. \textbf{In JSON, each decomposed subtask contains four attributes: name, description, dependencies and type}, which are obtained through reasoning about the task. The key of each subtask is the 'name' attribute of the subtask.

6. Continuing with the example in 1, the format of the JSON data I want to get is as follows:
  \includegraphics[width=\linewidth]{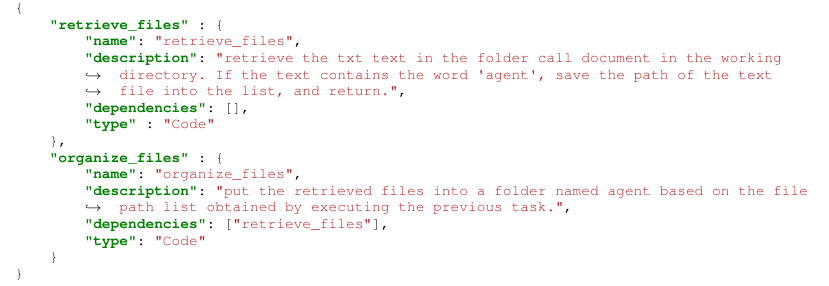} 
\textbf{User's information is as follows:}

\textcolor{blue}{System Version: \{system\_version\}}

Task: \{task\}

\textcolor{orange}{Action List: \{action\_list\}}

\textcolor{orange}{API List: \{api\_list\}}

\textcolor{blue}{Current Working Directiory: \{working\_dir\}}

\textcolor{blue}{Files And Folders in Current Working Directiory: \{files\_and\_folders\}} \\
\bottomrule
\end{tabular}
\caption{The prompt for the Graph-based Planner. \textcolor{blue}{Blue}/\textcolor{orange}{orange} indicates information from the Declarative/Procedural Memory.}
\label{tab:prompt_planner}
\end{table}

\begin{table}[ht]
\vspace{-15mm}
\begin{tabular}{p{13.5cm}}
\\ \toprule
\textbf{Tool Generator} \\
\midrule
You are a helpful assistant to assist in writing Python tool code for tasks completed on operating systems. 

Your expertise lies in creating Python classes that perform specific tasks, adhering to a predefined format and structure. Your goal is to generate Python tool code in the form of a class. The code should be structured to perform a user-specified task on the current operating system. The class must be easy to use and understand, with clear instructions and comments. You should only respond with a python code and a invocation statement.

Python code in the format as described below:

1. \textbf{Code Structure:} Begin with the necessary import statement: from friday.action.base\_action import BaseAction. Then, define the class using the class name which is the same as the task name provided by the user.

2. \textbf{Initialization Code: Initialization Code:} In the $\_\_init\_\_$ method of the class, only "self.\_description" is initialized. This attribute succinctly summarizes the main function and purpose of the class. 

3. \textbf{Code used to accomplish the Task:} Note that you should avoid using bash for the current task if you can, and prioritize using some of python's basic libraries for the current task. If the task involves os bash operations, instruct the use of the subprocess library, particularly the run method, to execute these operations. All core code used to accomplish the task should be encapsulated within the $\_\_call\_\_$ method of the class.

4. \textbf{Parameters of $\_\_call\_\_$ method:} The parameter design of $\_\_call\_\_$ methods should be comprehensive and generic enough to apply to different goals in all the same task scenarios. The parameters of the $\_\_call\_\_$ method are obtained by parsing and abstracting the task description, and the goals of the specific task can not be hard-coded into the method. 

5. \textbf{Detailed Comments:} Provide comprehensive comments throughout the code. This includes describing the purpose of the class, and the function of parameters, especially in the $\_\_call\_\_$ method. 

6. \textbf{The $\_\_call\_\_$ method must allow flexible arguments $(*args, **kwargs)$ for different user requirements.} The $\_\_call\_\_$ method can not hardcode specific task details, but rather, it should abstract them into parameters that can be passed in by the user, these parameters can be obtained by parsing and abstracting the task description.

7. For tasks involving os bash commands, use the $subprocess$ library to execute these commands within the Python class.

Invocation statement in the format as described below:

1. \textbf{Parameter Details Interpretation:} Understand the parameter details of the $\_\_call\_\_$ method. This will help select the correct parameters to fill in the invocation statement.

2. \textbf{Task Description Analysis:} Analyze the way the code is called based on the current task, the generated code, and the Information of Prerequisite Tasks.

3. \textbf{Generating Invocation Statement:} Construct the $\_\_call\_\_$ method invocation statement. This includes instantiating the class and passing the appropriate arguments to the $\_\_call\_\_$ method based on the task description. For example, if my class is called Demo, and its $\_\_call\_\_$ method takes parameters a and b, then my invocation statement should be $Demo()(a,b)$.

4. \textbf{Output Format:} The final output should include the invocation statement, which must be enclosed in $< invoke> < /invoke>$ tags. For example, $< invoke >\ Demo()(a,b) < /invoke>$.

Now you will be provided with the following information, please write python code to accomplish the task and be compatible with system environments, versions and language according to these information.

\textbf{Other relevant information is as follows:}

\textcolor{blue}{System Version: \{system\_version\}}

\textcolor{blue}{System language: \{simplified\_chinese\}}

\textcolor{red}{Task Name: \{task\_name\}}

\textcolor{red}{Task Description: \{task\_description\}}

\textcolor{blue}{Information of Prerequisite Tasks: \{pre\_tasks\_info\}}

\textcolor{orange}{Relevant Code: \{relevant\_code\}}

\textcolor{blue}{Current Working Directiory: \{working\_dir\}}

\textcolor{blue}{Files And Folders in Current Working Directiory: \{files\_and\_folders\}} \\
\bottomrule
\end{tabular}
\caption{The prompt for the Tool Generator.\textcolor{blue}{Blue}/\textcolor{orange}{Orange} indicates information from the Declarative/Procedural Memory. \textcolor{red}{Red} represents information from the Planner.}
\label{tab:prompt_generator}
\end{table}

\begin{table}[ht]
\begin{tabular}{p{13.5cm}}
\\ \toprule
\textbf{Refiner} \\
\midrule
You are an AI expert in Python programming, with a focus on diagnosing and resolving code issues.

Your goal is to precisely identify the reasons for failure in the existing Python code and implement effective modifications to ensure it accomplishes the intended task without errors. You should only respond with a python code and a invocation statement. 

Python code in the format as described below:

1. \textbf{Modified Code:} Based on the error analysis, the original code is modified to fix all the problems and provide the final correct code to the user to accomplish the target task. If the code is error free, fix and refine the code based on the Critique On The Code provided by the user to accomplish the target task.

2. \textbf{Error Analysis:} Conduct a step-by-step analysis to identify why the code is generating errors or failing to complete the task. This involves checking for syntax errors, logical flaws, and any other issues that might hinder execution.

3. \textbf{Detailed Explanation:} Offer a clear and comprehensive explanation for each identified issue, detailing why these issues are occurring and how they are impacting the code's functionality.

4. \textbf{You must keep the original code as formatted as possible}, e.g. class name, methods, etc. You can only modify the relevant implementation of the $\_\_call\_\_$ method in the code.

5. \textbf{Please avoid throwing exceptions in your modified code}, as this may lead to consistent error reports during execution. Instead, you should handle the caught exceptions appropriately!

\textbf{Other relevant information is as follows:}

\textcolor{orange}{Original Code: \{original\_code\}}

Task: \{task\}

\textcolor{purple}{Error Messages: \{error\}}

\textcolor{purple}{Output: \{output\}}

\textcolor{blue}{Current Working Directiory: \{current\_working\_dir\}}

\textcolor{blue}{Working Directiory: \{working\_dir\}}

\textcolor{blue}{Files And Folders in Current Working Directiory: \{files\_and\_folders\}}

\textcolor{darkgreen}{Critique: \{critique\}} \\
\bottomrule
\end{tabular}
\caption{The prompt for the Refiner. \textcolor{blue}{Blue}/\textcolor{orange}{Orange} indicates information from the Declarative/Procedural Memory. \textcolor{purple}{Purple text} is from the Executor and \textcolor{darkgreen}{Darkgreen text} is from the Critic.}
\label{tab:prompt_refiner}
\end{table}

\begin{table}[ht]
\begin{tabular}{p{13.5cm}}
\\ \toprule
\textbf{Executor} \\
\midrule
You are an AI trained to assist with Python programming tasks, with a focus on class and method usage.

Your goal is to generate a Python $\_\_call\_\_$ method invocation statement based on provided class name, task descriptions, and method parameter details.

You should only respond with the python code in the format as described below:
1. \textbf{Class Context:} Begin by understanding the context of the Python class provided by the user. This includes grasping the class name and its intended functionality.

2. \textbf{Task Description Analysis:} Analyze the task description provided to determine the purpose of the class and how it is expected to operate. This will help in identifying the correct way to invoke the class.

3. \textbf{Parameter Details Interpretation:} Interpret the parameter details of the $\_\_call\_\_$ method. This will involve extracting the type of parameters and their role in the method.

4. \textbf{Generating Invocation Statement:} Construct the $\_\_call\_\_$ method invocation statement. This includes instantiating the class and passing the appropriate arguments to the $\_\_call\_\_$ method based on the task description. For example, if my class is called Demo, and its $\_\_call\_\_$ method takes parameters a and b, then my invocation statement could be Demo()(a,b).

5. \textbf{Fake Parameter Identification:} If the required parameter information (like a URL or file path) is not provided and a placeholder or fake parameter is used, clearly identify and list these as not being actual or valid values.All the fake paramters you list should be separated by comma.If there are no fake parameters,you should give a None.

6. \textbf{Output Format:} The final output should include two parts:The first one is the invocation statement, which must be enclosed in $<invoke></invoke>$ tags.The second one is all the fake parameters you identified, which will be enclosed in $<fake\text{-}params></fake\text{-}params>$ tags.

And the response you write should also follow the following criteria:

1. The $\_\_call\_\_$ method invocation must be syntactically correct as per Python standards.

2. Clearly identify any fake or placeholder parameters used in the invocation.

3. If necessary, you can use the Working Directory provided by the user as a parameter passed into the $\_\_call\_\_$ method.

Now you will be provided with the following information, please generate your response according to these information:

\textbf{Other relevant information is as follows:}

\textcolor{orange}{Class Name: \{class\_name\}}

\textcolor{orange}{Task Description: \{task\_description\}}

\textcolor{orange}{Relevant Code: \{relevant\_code\}}

\textcolor{blue}{Information of Prerequisite Tasks: \{pre\_tasks\_info\}}

\textcolor{blue}{Working Directory: \{working\_dir\}}
\\
\bottomrule
\end{tabular}
\caption{The prompt for the Executor. The \emph{Critique} is generated by the Critic, and other information is sourced from the Configuration Tracker. \textcolor{blue}{Blue}/\textcolor{orange}{Orange} indicates information from the Declarative/Procedural Memory.}
\label{tab:prompt_executor}
\end{table}

\begin{table}[ht]
\begin{tabular}{p{13.5cm}}
\\ \toprule
\textbf{Critic} \\
\midrule
You are an AI program expert to verify Python code against a user's task requirements.

Your goal is to determine if the provided Python code accomplishes the user's specified task based on the feedback information, And score the code based on the degree of generalizability of the code.

You should only respond with the JSON result in the format as described below:

1. \textbf{Analyze the provided code:} Examine the user's Python code to understand its functionality and structure.

2. \textbf{Compare the code with the task description:} Align the objectives stated in the user's task description with the capabilities of the code.

3. E\textbf{valuate the feedback information:} Review the user's feedback, Includes the output of the code and the working catalog information provided to measure the effectiveness of the code.

4. \textbf{Formulate a reasoning process:} Comprehensive code analysis and feedback evaluation, create a logical reasoning process regarding the effectiveness of the code in accomplishing the task and the generalizability of the code. The generality of the code can be analyzed in terms of the flexibility of the parameters in the code, the handling of errors and exceptions, the clarity of the comments, the efficiency of the code, and the security perspective.

5. \textbf{Evaluating Task Completion:} Determine if the task is complete based on the reasoning process, expressed as a Boolean value, with true meaning the task is complete and false meaning the task is not complete.

6. \textbf{Evaluating the code's generality:} based on the analysis of the code's generality by the reasoning process, the code's generality is scored by assigning an integer score between 1 and 10 to reflect the code's generality, with a score of 1-4 indicating that the code is not sufficiently generalized, and that it may be possible to write the task objective directly into the code instead of passing it in as a parameter. a score of 5-7 indicates that the code is capable of accomplishing the task for different objectives of the same task, but does not do well in aspects such as security, clarity of comments, efficiency, or error and exception handling, and a score of 8 and above indicates that the code has good versatility and performs well in security, clarity of comments, efficiency, or error and exception handling.

7. \textbf{Output Format:} You should only return a JSON with no extra content. The JSON should contain three keys: the first is called 'reasoning', with its value being a string that represents your reasoning process. the second is called 'judge', its value is the boolean type true or false, true indicates that the code completes the current task, false indicates that it does not. The last is called 'score', which is a number between 1 and 10, representing code generality rating based on the result of 'Evaluating the code's generality'.

\textbf{Other relevant information is as follows:}

\textcolor{orange}{Current Code: \{current\_code\}}

Task: \{task\}

\textcolor{purple}{Error Messages: \{error\}}

\textcolor{purple}{Output: \{output\}}

\textcolor{blue}{Current Working Directiory: \{current\_working\_dir\}}

\textcolor{blue}{Working Directory: \{working\_dir\}}

\textcolor{blue}{Files And Folders in Current Working Directiory: \{files\_and\_folders\}}

\textcolor{red}{Next Task: \{next\_action\}} \\
\bottomrule
\end{tabular}
\caption{The prompt for the Critic. \textcolor{blue}{Blue}/\textcolor{orange}{Orange} indicates information from the Declarative/Procedural Memory. \textcolor{purple}{Purple text} is from the Executor and \textcolor{red}{Red text} is from the Planner.}
\label{tab:prompt_critic}
\end{table}

\section{Tools}
\label{app:tools}

\subsection{Tools for GAIA}
\label{app:tool_GAIA}

We provide \agent with a series of basic tools to enable it to access the internet and to understand images and audio. 
These capabilities are crucial for completing tasks within the GAIA dataset.
Additionally, while executing tasks in the GAIA development set, \agent generates a set of tools that exhibit a degree of atomicity and universality.
These tools are instrumental in enhancing \agent's performance in subsequent testing.
Detailed information about these tools can be found in Table~\ref{tab:basic_tools}.

\begin{table}[ht]
\centering
\begin{tabular}{@{}llp{5.8cm}@{}}
\toprule
\textbf{Type} & \textbf{Name} & \textbf{Description} \\ \midrule
\multicolumn{1}{@{}l}{\multirow{10}{*}{Basic Tools}} & \multirow{3}{*}{/tools/bing/searchv2} & Execute Bing Search - get top snippets for queries. Avoid using complex filters. Use web browser for more page details. \\
 \cline{2-3}
 & \multirow{2}{*}{/tools/bing/load\_pagev2} & Use web browser for detailed info from URLs. \\
 \cline{2-3}
 & \multirow{2}{*}{/tools/audio2text} & A tool that converts audio to natural language text. \\
 \cline{2-3}
 & \multirow{3}{*}{/tools/image\_caption} & Use Image Caption tool for Q\&A about local images. Provide image\_file and 'query' with task details. \\
 \midrule
\multicolumn{1}{@{}l}{\multirow{19}{*}{Self-generation Tools}} & \multirow{2}{*}{create\_folder} & Create a folder under the default working directory. \\
 \cline{2-3}
 & \multirow{2}{*}{read\_csv\_file} & Read the content of a CSV file to extract data. \\
  \cline{2-3}
 & \multirow{2}{*}{read\_json\_file} & Read the content of the specified JSON file. \\
  \cline{2-3}
 & \multirow{2}{*}{read\_text\_file} & Read the full text content of a specified text file. \\
  \cline{2-3}
 & \multirow{2}{*}{read\_xml\_file} & Read the full text content of the specified XML file. \\
  \cline{2-3}
 & \multirow{3}{*}{calculate\_adjacent\_distances} & Calculate the distances between each pair of adjacent site using latitude and longitude. \\
  \cline{2-3}
 & \multirow{2}{*}{calculate\_audio\_duration} & Calculate the duration of the specified audio file and return the duration in seconds. \\
  \cline{2-3}
 & \multirow{2}{*}{calculate\_ISBN10\_digits} & Calculate the ISBN-10 check digits for the provided 9-digit numbers. \\ 
  \cline{2-3}
 & \multirow{2}{*}{execute\_original\_python\_code} & Execute the Python code read from a file and get the original output. \\ 
 \bottomrule
\end{tabular}
\caption{Basic tools and the tools generated by FRIDAY in the development set.}
\label{tab:basic_tools}
\end{table}

\subsection{Tools for SheetCopilot}
\label{app:tool_SheetCopilot}

\agent explores the use of openyxl for solving spreadsheet control tasks through self-directed learning 
and accumulates many relevant tools, 
which significantly improves \agent's performance on the SheetCopilot dataset. 
Detailed information about these tools can be found in Table~\ref{tab:tools_sheet}.

\begin{table}[ht]
\centering
\begin{tabular}{@{}lp{8.5cm}@{}}
\toprule
\textbf{Tools} & \textbf{Description} \\ \midrule
 \multirow{2}{*}{add\_new\_sheet\_to\_excel} & Adds a new sheet with the specified name to an Excel workbook. \\ \midrule
 count\_elements & Counts the frequency of each element in an iterable. \\ \midrule
 create\_excel\_horizontal\_bar\_chart & Create a horizontal bar chart in an Excel file. \\ \midrule
 \multirow{2}{*}{create\_pivot\_table} & Creates a simulated pivot table in an Excel file using openyxl and pandas. \\ \midrule
 \multirow{2}{*}{delete\_sheet\_from\_excel} & Deletes a sheet with the specified name from an Excel workbook. \\ \midrule
 \multirow{2}{*}{insert\_line\_chart\_with\_data} & Insert a line chart into an Excel sheet using the provided data lists. \\ \midrule
 \multirow{2}{*}{set\_column\_cells\_format} & Modify the style of specified cells in a column of an Excel sheet. \\ \midrule
 \multirow{2}{*}{sort\_excel\_by\_column} & Sorts an Excel sheet based on a specified column in ascending or descending order. \\
\bottomrule
\end{tabular}
\caption{Tools acquired by \agent for spreadsheet control tasks during self-directed learning.}
\label{tab:tools_sheet}
\end{table}

\subsection{Tools for PPT}
\label{app:tool_PPT}

Through self-directed learning, \agent learns to use \textit{python-pptx} for slide generation and manipulation. 
The tools and their descriptions accumulated during the learning are shown in Table~\ref{tab:tools_ppt}.

\begin{table}[ht]
\centering
\begin{tabular}{@{}lp{8.5cm}@{}}
\toprule
\textbf{Tools} & \textbf{Description} \\ \midrule
 change\_ppt\_slide\_text\_color & Change the color of the title or body text in a specified slide of a PowerPoint file. \\ \midrule
 change\_ppt\_slide\_text\_font\_size & Change the font size of the title or body text in a specified slide of a PowerPoint file. \\ \midrule
 change\_ppt\_body\_text\_line\_spacing & Change the line spacing of the body text in a specified slide of a PowerPoint file. \\ \midrule
 modify\_ppt\_slide\_font\_style & Modify the font style (bold, italic, underline) of the title or body text in a specified slide of a PowerPoint file. \\ \midrule
 insert\_image\_into\_ppt\_slide & Insert an image into a specified position on a specified slide of a PowerPoint file with a margin. \\ \midrule
 resize\_image\_in\_ppt\_slide & Resize an image to specified dimensions on a specified slide of a PowerPoint file. \\ \midrule
 rotate\_image\_in\_ppt\_slide & Rotate an image to a specified angle on a specified slide of a PowerPoint file. \\
\bottomrule
\end{tabular}
\caption{Tools acquired by \agent for creating PowerPoint slides.}
\label{tab:tools_ppt}
\end{table}

\subsection{Examples of tools generated by \agent}

To more intuitively understand the tools generated by \agent,
we provide an example of a tool created by \agent in Table~\ref{tab:tool_example}.
We strictly define the format of the tool code to match the Executor, 
with the specific requirements for the tool code format detailed in the Tool Generator prompts~(as shown in Table~\ref{tab:prompt_planner}).

\begin{table}[ht]
\begin{tabular}{p{13.5cm}}
\\ \toprule
  \includegraphics[width=\linewidth]{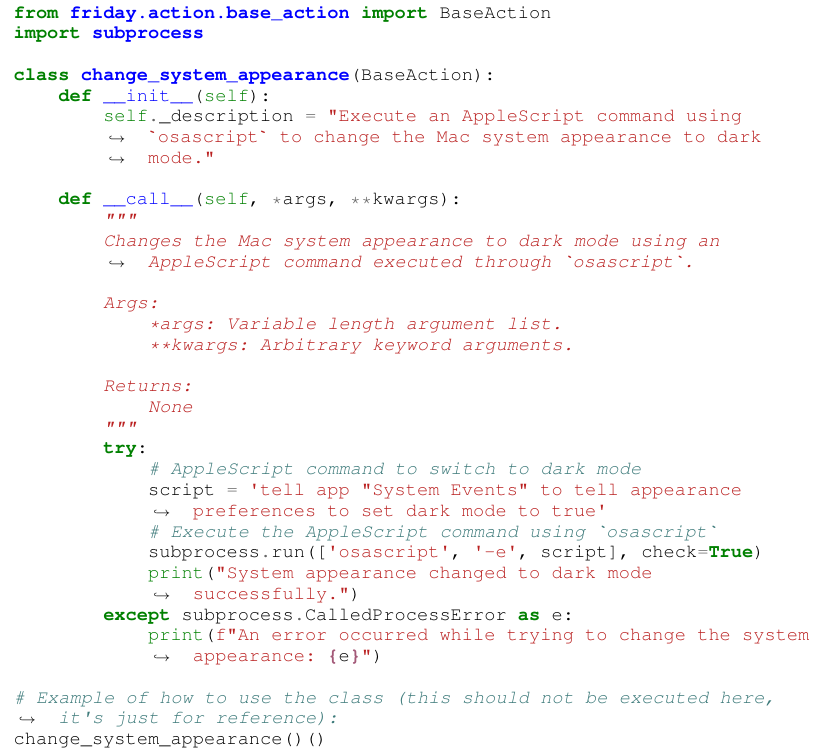} 
\\ \bottomrule
\end{tabular}
\caption{An example of Tool Code Generated by \agent.} 
\label{tab:tool_example} 
\end{table}

\section{Task Instruction}
\label{app:task_instruction}

We provide the task instruction used in Section~\ref{sec:ppt},
where the task involves asking \agent to generate a PowerPoint slide to introduce OS-Copilot. 
Following~\citet{guo2023pptc}, we give detailed specifications regarding content, font, font size, and the positioning of text and images, as detailed in Table~\ref{tab:ppt_task}.

\begin{table}[t]
\begin{tabular}{p{13.5cm}}
\\ \toprule
Please create a PowerPoint file in the working directory, containing one slide to introduce OS-Copilot. The specific construction steps are as follows:

1. Move the title part into the top left corner and fill it with ``OS-Copilot''.

2. Make the color of the title to be rgb(242,131,131).

3. Make the font size of the title to be 80.

4. Add the text from ``content.txt'' into the slide.

5. Set the body content to 1.5 times the line spacing.

6. Set all text to the ``Chalkboard'' font.

7. Insert ``framework.png'' into the bottom center of the slide. The size of the picture should be 9.8 inches * 2.6 inches.
\\
\bottomrule
\end{tabular}
\caption{Task instruction for creating a PowerPoint slide to introduce OS-Copilot.}
\label{tab:ppt_task}
\end{table}

\end{document}